\documentclass{bmvc2k}

\usepackage{amsfonts}  
\usepackage{amsmath}
\usepackage{amssymb}
\usepackage{booktabs}
\usepackage[capitalize]{cleveref}
\usepackage{wrapfig}
\def\cf{\emph{cf}\bmvaOneDot}

\crefname{table}{Tab.}{Tabs.}
\Crefname{table}{Tab.}{Tabs.}
\crefname{section}{Sec.}{Secs.}
\Crefname{section}{Sec.}{Secs.}
\crefname{appendix}{App.}{Apps.}
\Crefname{appendix}{App.}{Apps.}

\title{TryOffDiff: Virtual-Try-Off via High-Fidelity Garment Reconstruction using Diffusion Models}

\addauthor{Riza Velioglu}{rvelioglu+bmvc@techfak.de}{1}
\addauthor{Petra Bevandić}{pbevandic+bmvc@techfak.de}{1}
\addauthor{Robin Chan}{chan+bmvc@tu-berlin.de}{2}
\addauthor{Barbara Hammer}{bhammer+bmvc@techfak.de}{1}

\addinstitution{
 Machine Learning Group, CITEC\\
 Bielefeld University\\
 Germany
}
\addinstitution{
 Institute of Mathematics\\
 Technical University Berlin\\
 Germany
}

\runninghead{Velioglu \etal}{TryOffDiff}

\def\eg{\emph{e.g}\bmvaOneDot}

\def\etal{\emph{et al}\bmvaOneDot}

\begin{document}
\maketitle

\begin{center}
    \captionsetup{type=figure}
        \includegraphics[width=\textwidth]{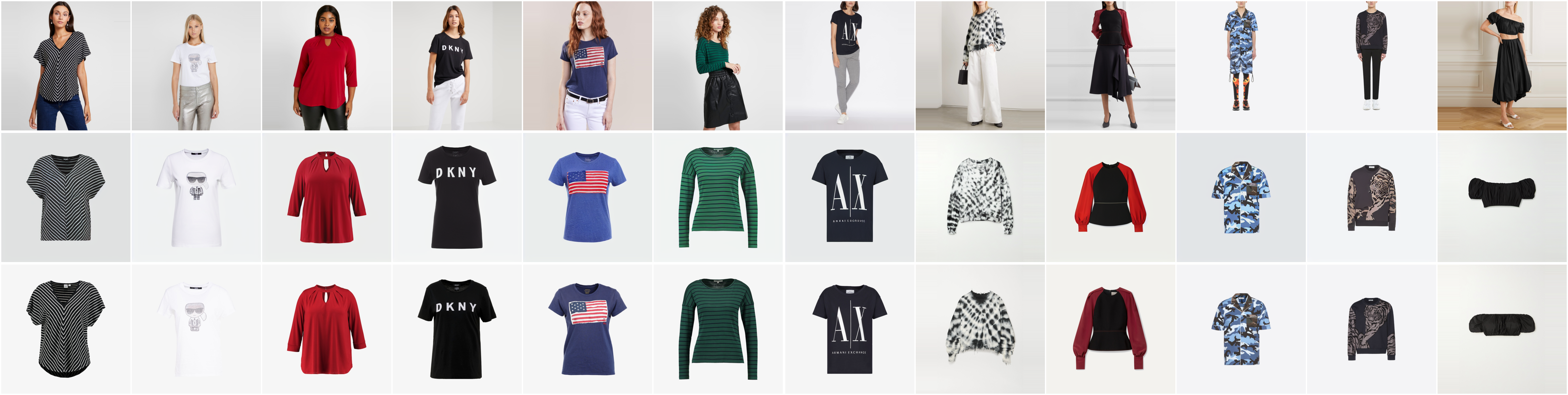}
    \captionof{figure}{\textbf{Virtual Try-Off results from our method}. 
    Each row shows the input, prediction, and ground truth for samples from VITON-HD (cols 1-6) and Dress Code (cols 7-12). TryOffDiff synthesizes garments on a clean background in a standard pose, accurately preserving fine details like patterns and logos from a single reference image.
    }
    \label{fig:cover}
\end{center}%

\begin{abstract}
This paper introduces Virtual Try-Off (VTOFF), a novel task generating standardized garment images from single photos of clothed individuals. 
Unlike Virtual Try-On (VTON), which digitally dresses models, VTOFF extracts canonical garment images, demanding precise reconstruction of shape, texture, and complex patterns, enabling robust evaluation of generative model fidelity. 
We propose TryOffDiff, adapting Stable Diffusion with SigLIP-based visual conditioning to deliver high-fidelity reconstructions. 
Experiments on VITON-HD and Dress Code datasets show that TryOffDiff outperforms adapted pose transfer and VTON baselines.
We observe that traditional metrics such as SSIM inadequately reflect reconstruction quality, prompting our use of DISTS for reliable assessment.
Our findings highlight VTOFF’s potential to improve e-commerce product imagery, advance generative model evaluation, and guide future research on high-fidelity reconstruction.
Demo, code, and models are available at: \small \url{https://rizavelioglu.github.io/tryoffdiff}

\end{abstract}

\section{Introduction}\label{sec:introduction}

High-quality, standardized product images are vital in e-commerce, particularly in fashion, where visuals drive consumer purchasing decisions~\cite{xia2020creating,VANDERHEIDE2013570}. Producing such images is resource-intensive, requiring specialized equipment and extensive post-processing. 
While existing approaches like Virtual Try-On (VTON)~\cite{jetchev2017conditional} focus on enhancing customer experience by synthesizing images of people wearing selected garments—using a garment image and a reference person image—they do not address the creation of the garment images themselves. VTON advances human-centric generation, involving pose transfer, garment warping, and texture preservation\cite{han2018viton}, but assumes access to clean product shots.
To address this gap, we introduce Virtual Try-Off (VTOFF): the task of extracting standardized garment images from photos of clothed individuals. To date, this inverse task has received little attention and has yet to be formally defined.

\begin{wrapfigure}{R}{0.43\textwidth}
    \vspace*{-\baselineskip} 
    \centering
    \includegraphics[width=\linewidth]{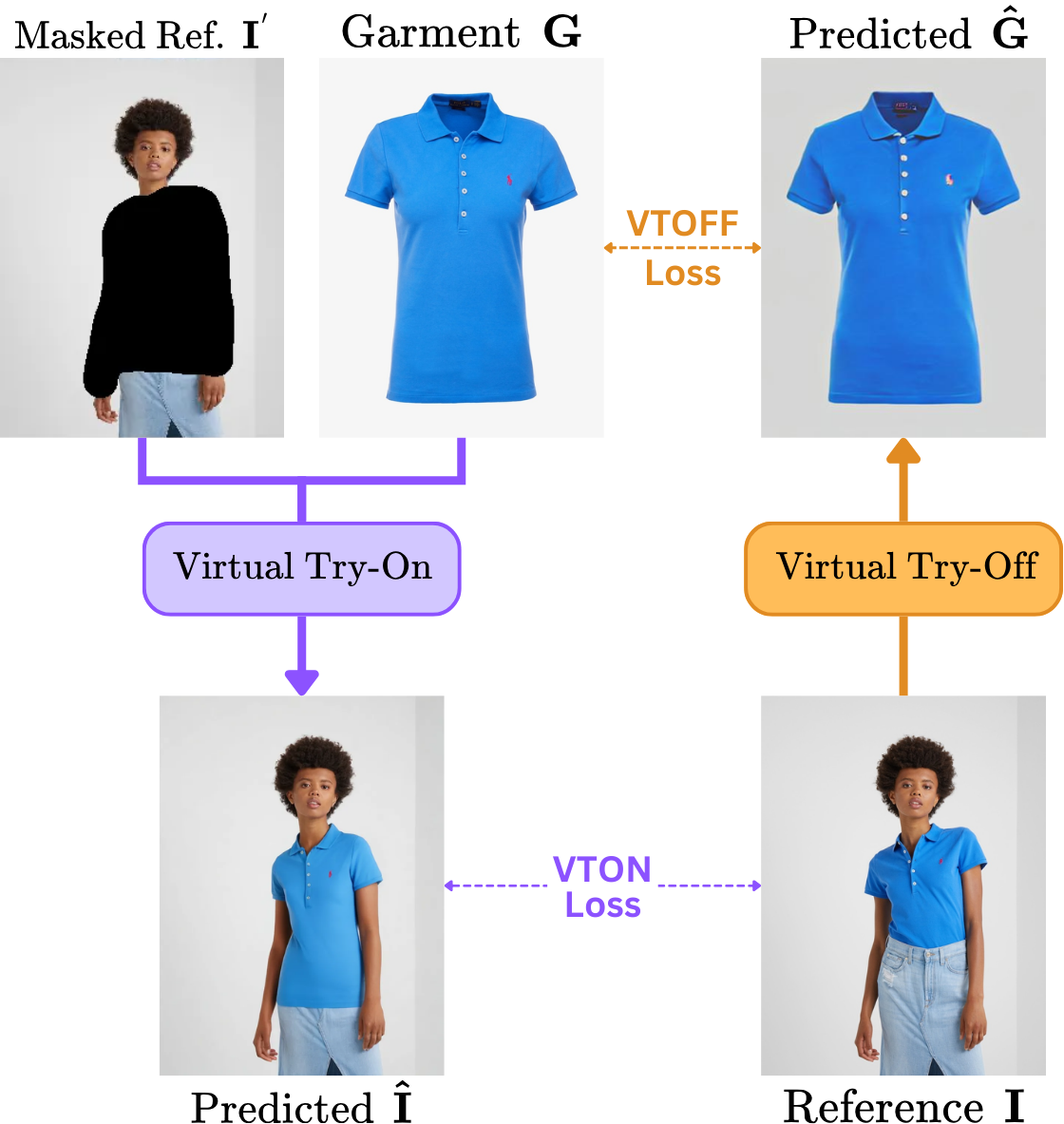}
    \caption{\textbf{VTON vs. VTOFF.}
    Left: Virtual Try-On generates a dressed person from a garment image and a masked reference.
    Right: Virtual Try-Off reconstructs a canonical garment form from a photo of a clothed person.
    The two tasks form a cycle, where one's output can serve as other's input.
    }
    \label{fig:vtoff}
\end{wrapfigure}

VTOFF addresses critical needs across multiple domains. In e-commerce, it offers a practical solution for generating catalog-ready garment images from diverse real-world photos. This reduces reliance on costly photography setups and manual editing, empowering smaller vendors to access professional-quality visuals. It supports customer-to-product retrieval by providing consistent garment images for recommendation systems~\cite{DeepFashion2}. Environmentally, VTOFF enhances visual clarity of garment details, aligning customer expectations with product appearance and potentially reducing product returns, which are a major contributor to the fashion industry's environmental footprint~\cite{dzyabura2023leveraging,hartmann2024power}. From a research perspective, VTOFF presents several complex challenges that advance the field of high-fidelity garment reconstruction. These include handling occlusions, deformations, and varying poses, as well as managing geometric and appearance transformations while preserving fine-grained details such as textures, patterns, and logos. The diversity of real-world photos, which vary in background, lighting, and camera quality, further introduces unique hurdles in domain adaptation and robust feature extraction, making VTOFF a compelling area for innovation.
Moreover, VTON and VTOFF exist in a complementary relationship, as illustrated in~\cref{fig:vtoff}. This cyclic connection enables: (1) improved training through consistency constraints in the loss function~\cite{wang2024fldm}, (2) generation of synthetic data for both tasks~\cite{chong2024catvton,shen2025mfp}. This synergy boosts model robustness and opens new avenues for fashion image generation.

A key advantage of VTOFF over VTON is its evaluation clarity. VTON suffers from ill-defined target outputs, which complicates quality assessment. Generated images often exhibit stylistic variations, such as garments appearing tucked, untucked, or altered in fit (see \cref{fig:vtoff}), introducing plausible yet inconsistent results that obscure true garment fidelity~\cite{theis2016note}. 
Existing evaluation metrics, primarily designed for broad generative quality~\cite{heusel2017gans}, are sensitive to irrelevant regions like backgrounds and fail to focus on garment-specific details, making it hard to identify performance issues~\cite{ding2020image,tang2011learning}. In contrast, VTOFF provides a \emph{clear and precise target}: the standardized garment image. This well-defined output enables more accurate assessment of reconstruction quality, particularly in capturing shape, texture, and patterns, making VTOFF an ideal task for evaluating generative models in fashion contexts.
By bridging e-commerce solutions and research opportunities, VTOFF advances fashion technology. 
This paper lays the groundwork for its exploration.
Our main contributions are:
\begin{itemize}
    \item We introduce \textbf{VTOFF}, a novel task to generate standardized garment images from photos of clothed individuals, addressing e-commerce needs and research challenges.
    \item We leverage VTOFF's well-defined targets to expose the limitations of conventional metrics and validate DISTS as a more reliable alternative.
    \item We propose \textbf{TryOffDiff}, a framework that adapts pretrained diffusion models with pretrained image encoders for high-fidelity garment reconstruction.
    \item Experiments on the VITON-HD and Dress Code datasets show TryOffDiff outperforms adapted baselines with minimal pre- and post-processing.
\end{itemize}

\section{Related Work}\label{sec:related_work}

This section reviews prior research relevant to the proposed task and method, covering garment reconstruction, image-based virtual try-on, view synthesis, pose transfer, and conditional diffusion models.

\textbf{Garment Generation}
is central to VTOFF. Early work, such as TileGAN~\cite{zeng2020tilegan}, tackled VTOFF using a two-stage GAN-based approach. While innovative for its time, TileGAN did not formally define VTOFF as a standalone problem, and its reliance on GANs has been eclipsed by diffusion models. Limited evaluation, restricted datasets, and unavailable code likely curbed its adoption, highlighting the need for a more comprehensive framework.
Later efforts in garment synthesis deviate from VTOFF's image-only focus. ARMANI~\cite{zhang2022armani} and DressCode~\cite{he2024dresscode} emphasize text-to-garment generation, producing garments (including 3D models in the latter) from textual descriptions. Similarly, SGDiff~\cite{sun2023sgdiff} and DiffCloth~\cite{zhang2023diffcloth} leverage diffusion models—finetuning GLIDE, and Stable Diffusion, respectively—to synthesize garments, but they rely on multi-modal inputs like text combined with style images or sketches. These approaches, while advancing synthesis, sidestep VTOFF's core challenge of reconstructing garments from real-world photos alone.
Closer to VTOFF, FLDM-VTON~\cite{wang2024fldm} incorporates a VTOFF-like component in its loss function to improve VTON training. However, it does not address VTOFF explicitly, limiting its relevance to our problem definition.
As a result, prior work has overlooked the specific challenges of VTOFF, including occlusions, garment deformations, and diverse real-world photo conditions. 
To the best of our knowledge, we are the first to formally define VTOFF, explore its potential, and propose appropriate evaluation protocols. 

\textbf{Image-based Virtual Try-On}
aims to generate realistic images of a person wearing a specified garment, preserving their identity, pose, and body shape while capturing fine garment details. Early efforts, such as CAGAN\cite{jetchev2017conditional} introduced the task using a cycle-GAN approach, while VITON\cite{han2018viton} formalized it as a two-step supervised framework: warping the garment and blending it onto the person. CP-VTON\cite{wang2018toward} refined warping with a learnable thin-plate spline transformation, later enhanced by dense flow\cite{han2019clothflow} and appearance flow~\cite{ge2021parser} for better pixel-level alignment. Despite progress, challenges with complex textures remain.
Recent developments have shifted toward GAN-based and diffusion-based methods. FW-GAN\cite{dong2019fw} and PASTA-GAN\cite{xie2021towards} leveraged GANs, but their instability prompted a move to diffusion models, such as IDM-VTON\cite{choi2024improving}, OOTDiffusion\cite{xu2024ootdiffusion}, and CatVTON\cite{chong2024catvton}, which provide greater stability and detail retention.
Notably, VTON and VTOFF differ fundamentally despite both involving garment manipulation. VTON models work with complete garment details, focusing on warping them to fit a target pose. Conversely, VTOFF models rely on partial, often occluded garment views from real-world images, requiring them to infer and reconstruct missing details from limited visual information.

\textbf{Image-based Novel View Synthesis \& Pose Transfer.}
NVS aims to generate realistic images from unseen viewpoints. 
Early approaches demanded extensive image sets per instance~\cite{kulkarni2015deep,tatarchenko2016multi,zhou2016view,sun2018multi,zhao2018multi}, 
whereas recent methods achieve synthesis from sparse views~\cite{tochilkin2024triposr,jang2024nvist}. However, NVS lacks the ability to handle pose variations, limiting its use in garment reconstruction. Pose transfer extends view synthesis by enabling pose alterations and inferring occluded regions. DiOr~\cite{cui2021dressing} employs a recurrent framework to sequentially dress subjects, while \cite{roy2023multi} leverages a GAN-based approach for enhanced pose transfer. Advanced methods like DreamPose~\cite{karras2023dreampose} and PoCoLD~\cite{han2023controllable} utilize diffusion models for pose-guided image and video synthesis. ViscoNet~\cite{cheong2024visconet} and PCDM~\cite{shen2023advancing} further improve control and detail fidelity in pose-driven synthesis.
Unlike pose transfer, which preserves scene attributes like lighting, background and subject appearance, VTOFF demands specific standards, such as consistent views, uniform sizing, and e-commerce catalog styling.

\textbf{Conditional Diffusion Models.}
Latent Diffusion Models~\cite{rombach2022high}~(LDMs) excel in synthesis, leveraging cross-attention~\cite{vaswani2017attention} for precise conditioning on text~\cite{betker2023improving,baldridge2024imagen,esser2024scaling} 
or images~\cite{saharia2022palette,parmar2023zero,saharia2022image}. 
Text-guided extensions like ControlNet~\cite{zhang2023adding} and T2I-Adapter~\cite{mou2024t2i} add spatial accuracy, while IP-Adapter~\cite{ye2023ipadapter} separates text and image features for flexibility. 
Despite the advancements, applying these models directly to garment reconstruction poses significant challenges: text-guided approaches demand impractically detailed prompts for each sample to specify product attributes, 
while existing image-guided models lack specialized mechanisms needed to meet the strict requirements of standardized product photography, such as precise alignment, and detail preservation.

\section{Methodology}\label{sec:methodology}

We formalize the virtual try-off task, propose an evaluation framework with suitable metrics, and detail the TryOffDiff model architecture.

\subsection{Virtual Try-Off}\label{sec:vtoff}
\textbf{Problem Formulation.}
Let $\mathbf{I} \in \mathbb{R}^{H \times W \times 3}$ be an RGB image of a clothed person, with height $H \in \mathbb{N}$ and width $W \in \mathbb{N}$. 
VTOFF aims to generate a standardized garment image $\mathbf{G} \in \{0, \ldots, 255\}^{H \times W \times 3}$ from $\mathbf{I}$, adhering to commercial catalog standards.
Formally, we seek to train a generative model to approximate the conditional distribution $P(G|C)$, where $G$ and $C$ denote garment images and reference images~(serving as condition), respectively.
Given a reference image $\mathbf{I}$, the model should produce a sample $\hat{\mathbf{G}} \sim Q(G | C = \mathbf{I})$ closely matching a ground-truth garment image $\mathbf{G} \sim P(G | C = \mathbf{I})$.

\noindent \textbf{Datasets.}
VTOFF does not require new data collection. It leverages existing VTON datasets such as VITON-HD~\cite{choi2021viton} and Dress Code~\cite{morelli2022dress}. These datasets provide input-target pairs $(\mathbf{I}, \mathbf{G})$ for training and evaluation, where $\mathbf{I}$ is an image of a clothed person (half-body or full-body) and $\mathbf{G}$ is the corresponding standardized garment image.

\begin{figure}[t]
  \centering
  \begin{subfigure}{0.15\linewidth}
    \includegraphics[width=0.99\linewidth]{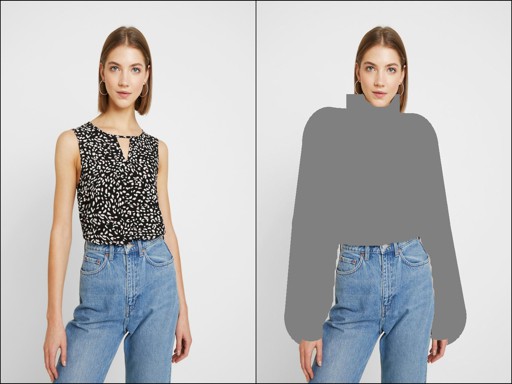}
    \caption{$82.4 \text{ / } 20.6$}
    \label{fig:ssim-a}
  \end{subfigure}
  \hfill
  \begin{subfigure}{0.15\linewidth}
    \includegraphics[width=0.99\linewidth]{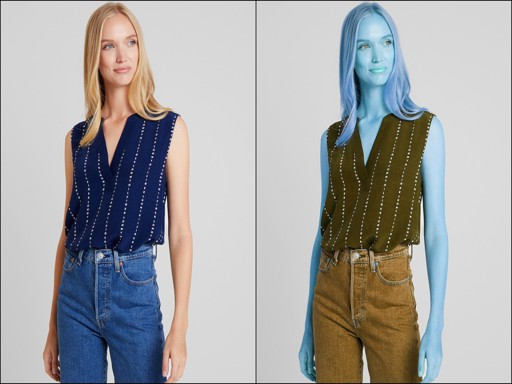}
    \caption{$96.8 \text{ / } 17.9$}
    \label{fig:ssim-b}
  \end{subfigure}
\hfill
  \begin{subfigure}{0.15\linewidth}
    \includegraphics[width=0.99\linewidth]{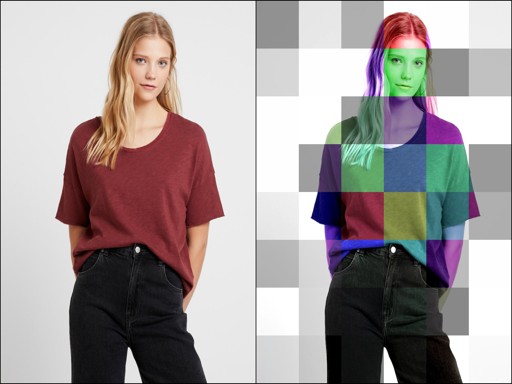}
    \caption{$88.3 \text{ / } 20.3$}
    \label{fig:ssim-c}
  \end{subfigure}
\hfill
  \begin{subfigure}{0.15\linewidth}
    \includegraphics[width=0.99\linewidth]{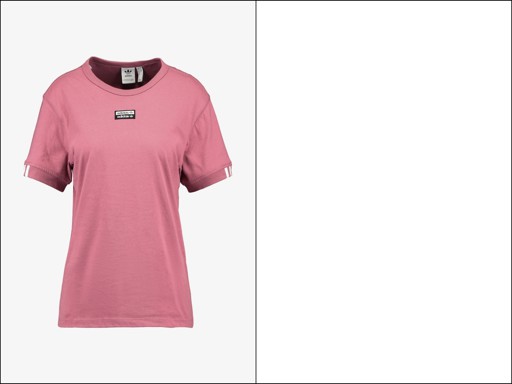}
    \caption{$86.0 \text{ / } 70.3$}
    \label{fig:ssim-d}
  \end{subfigure}
\hfill
  \begin{subfigure}{0.15\linewidth}
    \includegraphics[width=0.99\linewidth]{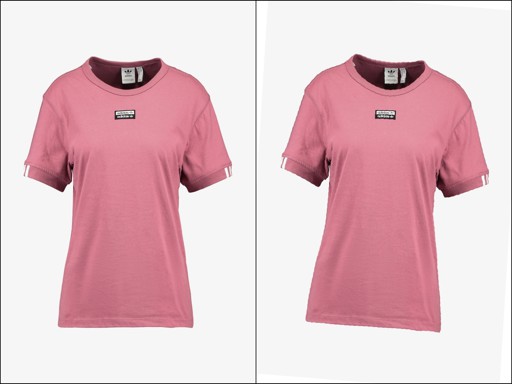}
    \caption{$75.0 \text{ / } 8.2$}
    \label{fig:ssim-e}
  \end{subfigure}
\hfill
  \begin{subfigure}{0.15\linewidth}
    \includegraphics[width=0.99\linewidth]{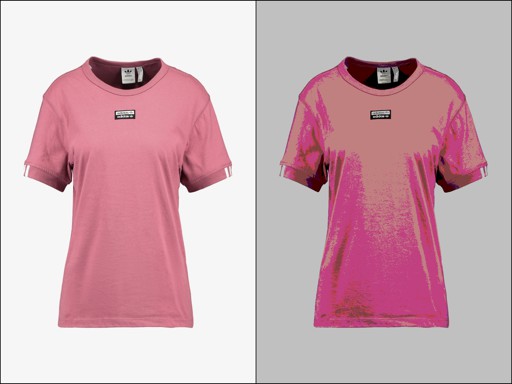}
    \caption{$86.4 \text{ / } 24.7$}
    \label{fig:ssim-f}
  \end{subfigure}  
  \\
  \caption{\textbf{Metric suitability for VTON and VTOFF (SSIM\textuparrow ~/~DISTS\textdownarrow).}
  A reference image is compared to:
  \textbf{(a)} masked-out garment,
  \textbf{(b)} hue-jittered,
  \textbf{(c)} patch-wise color-jittered image;
  and a garment image is compared to:
  \textbf{(d)} plain white,
  \textbf{(e)} slightly rotated,
  \textbf{(f)} randomly posterized image. 
  SSIM fails to penalize distortions, while DISTS better reflects judgment.
}
  \label{fig:metrics_failures}
\end{figure}

\noindent \textbf{Performance Measures.} 
Effective evaluation requires metrics that assess both reconstruction quality (pixel-level fidelity between $\hat{\mathbf{G}}$ and $\mathbf{G}$) and perceptual quality (visual realism per human judgment).
Full-reference metrics like \emph{Structural Similarity Index Measure} (SSIM) \cite{wang2004image} and its \emph{multi-scale} (MS-SSIM) 
and \emph{complex-wavelet}~(CW-SSIM) variants measure reconstruction but poorly align with human perception, as shown in prior studies~\cite{ding2020image, tang2011learning} and our own experiments~(\cref{fig:metrics_failures}).
For instance, \cref{fig:ssim-d} compares a garment image to a plain white image and still yields a high SSIM score (86/100), even higher than the garment image itself slightly rotated (\cref{fig:ssim-e}), highlighting a key failure mode. Indeed, we evaluated these metrics across the entire dataset, with results in~\cref{app:tab:eval_metrics_test} (appendix), confirming DISTS's robustness over traditional metrics for VTOFF.
Perceptual quality may be captured with no-reference metrics like \emph{Fréchet Inception Distance} (FID)~\cite{heusel2017gans} and \emph{Kernel Inception Distance} (KID)~\cite{binkowski2018demystifying}, as they compare feature distributions. However, these are sensitive to sample size and outliers, making them unsuitable for single-image evaluation. Additionally, their reliance on Inception features \cite{szegedy2015going} misaligns with human judgment in assessing perceptual quality \cite{stein2024exposing}.
Given these limitations, we adopt \emph{Deep Image Structure and Texture Similarity} (DISTS) \cite{ding2020image} metric as our primary measure. DISTS compares images using VGG features~\cite{simonyan2015vgg}, combining low-level structural and high-level textural features through a weighting  scheme optimized via human-rated similarity. This yields a perceptual metric that better captures the quality of garment reconstructions in VTOFF.

\subsection{TryOffDiff} \label{sec:TryOffDiff}

\begin{figure}[t]
    \centering
    \includegraphics[width=\linewidth]{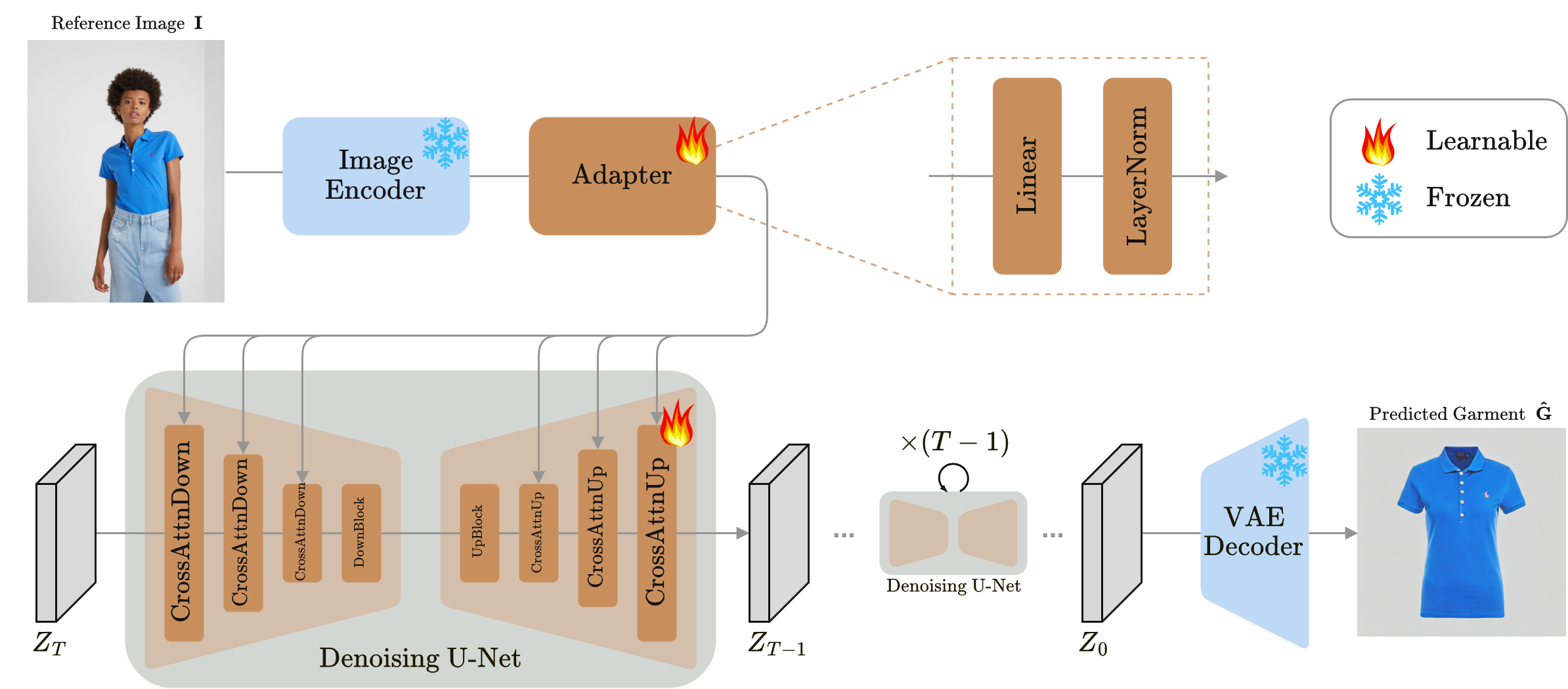}
    \caption{\textbf{TryOffDiff architecture.} SigLIP~\cite{zhai2023sigmoid} extracts features from a reference image, processed by adapter modules and embedded into pretrained Stable Diffusion v1.4 \cite{rombach2022high} by replacing text features in cross-attention layers, enabling image-guided generation. Joint training of adapter layers and diffusion model enables effective garment transformation.
    }
    \label{fig:TryOffDiff}
\end{figure}

TryOffDiff adapts Stable Diffusion \cite{rombach2022high} (v1.4), 
a latent diffusion model originally 
designed for text-conditioned image generation 
using CLIP~\cite{radford2021learning}, to perform image-guided generation by replacing text prompts with image features.
A core challenge in image-guided generation
is effectively incorporating visual features
into the conditioning mechanism of the generative model.
CLIP's ViT~\cite{radford2021learning} has become 
a popular choice for image feature extraction 
due to its general-purpose capabilities and joint embedding space for text and images.
Recently, SigLIP~\cite{zhai2023sigmoid} introduced modifications that improve performance, 
particularly for tasks requiring more detailed and domain-specific visual representations. 
Therefore, we use the SigLIP model as image feature extractor and
retain the entire sequence of token representations in its final layer
to preserve spatial information,
which we find essential for the capture of
fine-grained visual details 
and accurate garment reconstruction.
%
For an input image $\mathbf{I}$, the condition is derived as 
$\mathbf{C}(\mathbf{I}) = (\mathrm{LN} \circ \mathrm{Linear} \circ \mathrm{SigLIP})(\mathbf{I}) \in \mathbb{R}^{n \times m}$,
where SigLIP embeddings are linearly projected followed by layer normalization~(LN) \cite{ba2016layer}, \cf \cref{fig:TryOffDiff}.
These features are integrated into the denoising U-Net via cross-attention, with keys 
$\mathbf{K} = \mathbf{C}(\mathbf{I}) \cdot \mathbf{W}_k \in \mathbb{R}^{n \times d_k}$ and values 
$\mathbf{V} = \mathbf{C}(\mathbf{I}) \cdot \mathbf{W}_V \in \mathbb{R}^{n \times d_v}$,
where $\mathbf{W}_k \in \mathbb{R}^{m \times d_k}$ 
and $\mathbf{W}_v \in \mathbb{R}^{m \times d_v}$.
This conditions the denoising process on the visual features of reference image $\mathbf{I}$, enhancing alignment in the output.
We train adapter modules and finetune denoising U-Net, keeping SigLIP and VAE encoder-decoder frozen to leverage pretrained capabilities.


\begin{figure}[t]
  \centering
  \begin{subfigure}{0.49\textwidth}
    \includegraphics[width=\linewidth]{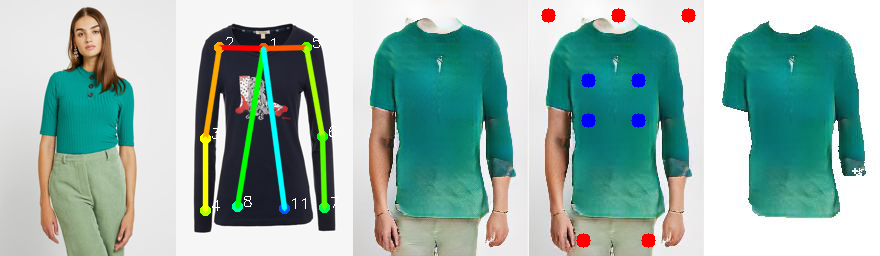}
    \caption{Left to right: reference image, fixed pose heatmap derived from target image, initial model output, SAM prompts, and final processed output.}
    \label{fig:baselines-pose}
  \end{subfigure}
  \hfill
  \begin{subfigure}{0.49\textwidth}
    \includegraphics[width=\linewidth]{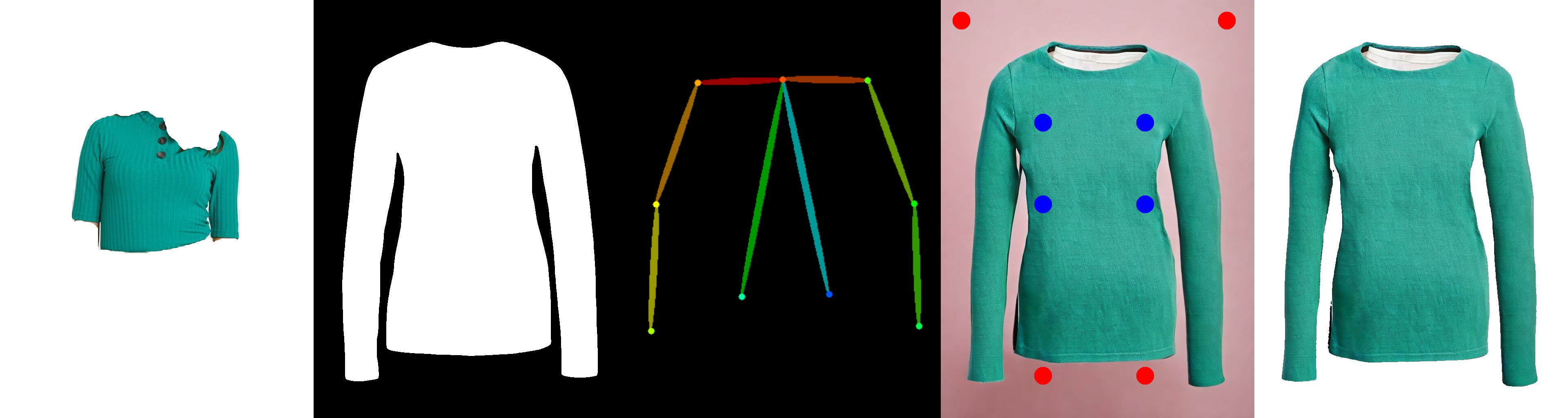}
    \caption{Left to right: masked conditioning image, mask image, pose image, initial model output with SAM prompts, and final processed output.}
    \label{fig:baselines-visconet}
  \end{subfigure}
  \hfill
  \begin{subfigure}{0.32\textwidth}
    \includegraphics[width=\linewidth]{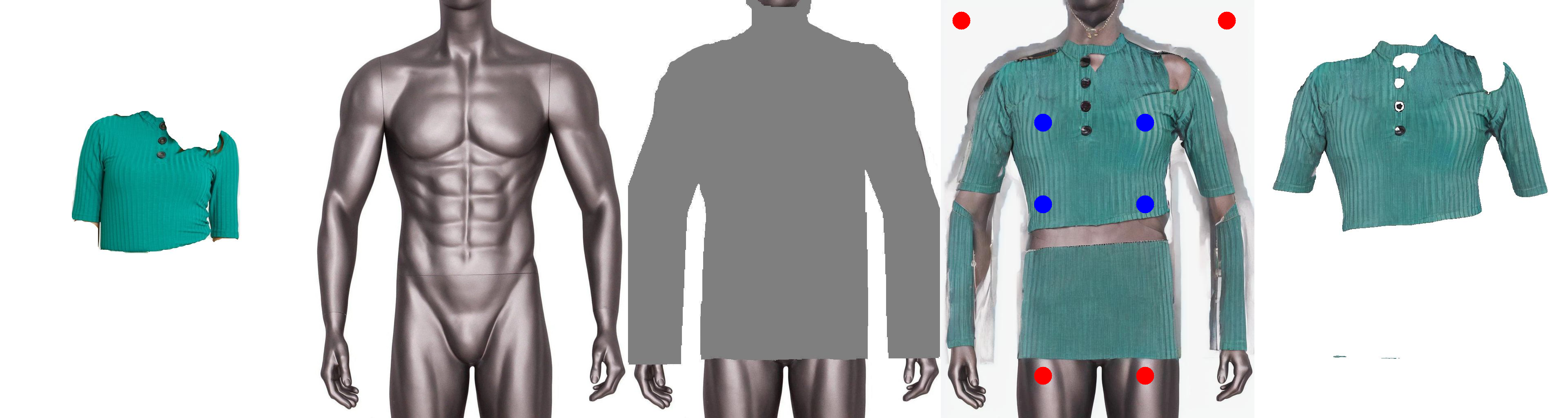}
    \caption{Left to right: masked garment, model image, masked model, initial model output with SAM prompts, and final output.}
    \label{fig:baselines-ootd}
  \end{subfigure}
  \hfill
  \begin{subfigure}{0.32\textwidth}
    \includegraphics[width=\linewidth]{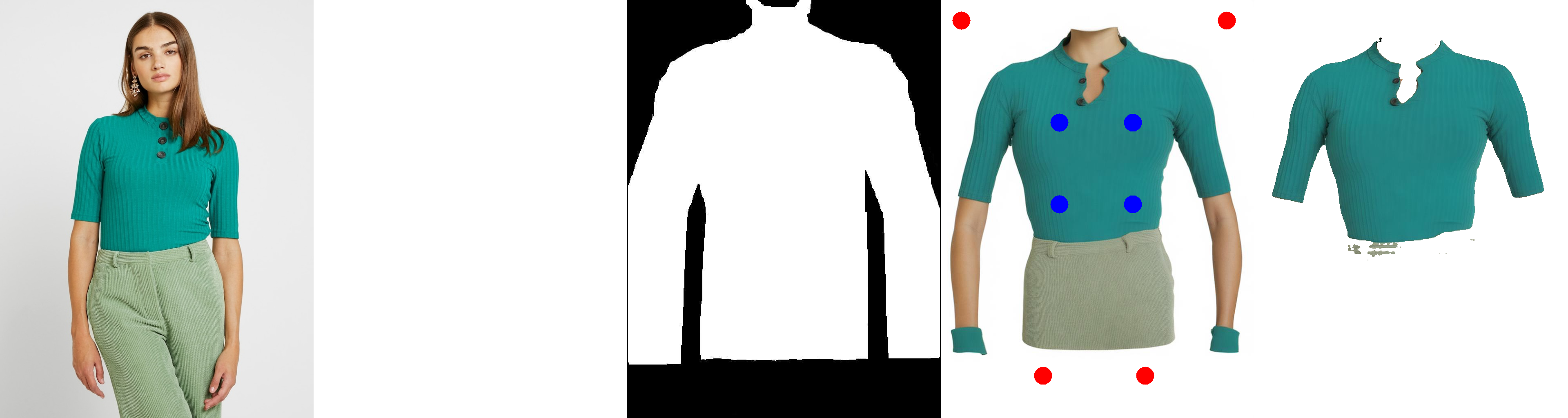}
    \caption{Left to right: conditioning garment, blank model, mask image, initial output with SAM prompts, and final output.}
    \label{fig:baselines-catvton}
  \end{subfigure}
  \hfill
  \begin{subfigure}{0.32\textwidth}
    \includegraphics[width=\linewidth]{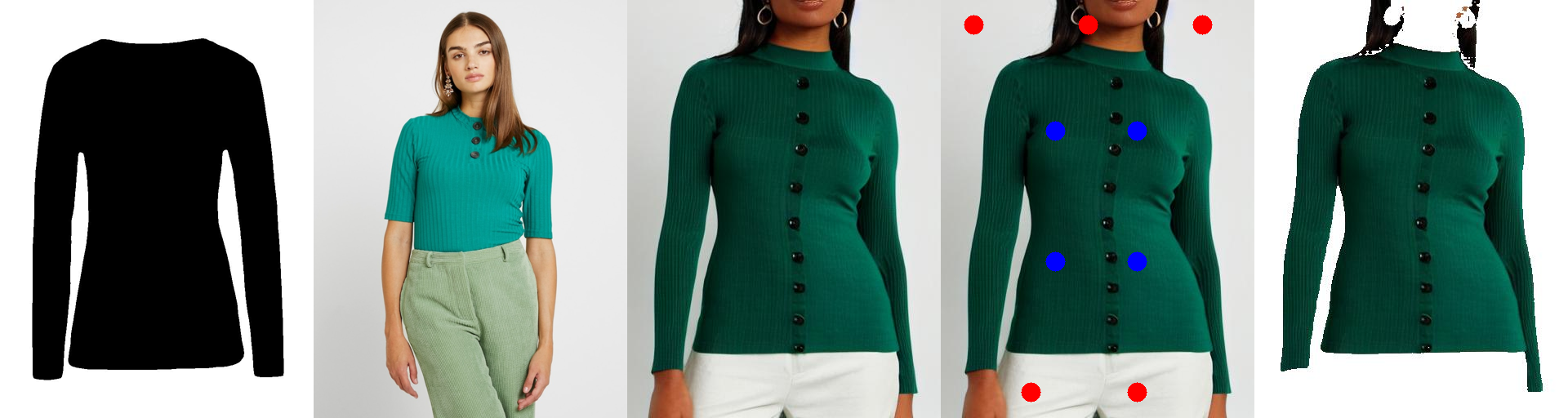}
    \caption{Left to right: starting image, image prompt, initial output, output with SAM prompts, and final output.}
    \label{fig:baselines-ip}
  \end{subfigure}  
  \caption{\textbf{Adapting baselines for VTOFF.} (a) GAN-Pose~\cite{roy2023multi} uses pose transfer; (b) ViscoNet~\cite{cheong2024visconet} employs view synthesis; (c) OOTDiffusion~\cite{xu2024ootdiffusion} and (d) CatVTON~\cite{chong2024catvton} are recent virtual try-on methods; (e) IP-Adapter~\cite{ye2023ipadapter} is a general-purpose image-conditioned baseline.}
  \label{fig:baselines}
\end{figure}
\section{Experiments}\label{sec:experiments}
We evaluate our proposed TryOffDiff model on the VTOFF task against a range of adapted baselines, including virtual try-on methods, pose transfer approaches, and a general-purpose image-conditioning method. 
We describe experimental setup for reproducibility, report quantitative results, and present qualitative comparisons to demonstrate TryOffDiff’s ability to reconstruct fine-grained garment details across diverse inputs.

\noindent \textbf{Datasets.}
We conduct experiments on VITON-HD~\cite{choi2021viton}, containing 13,679 high-resolution ($1024 \times 768$) pairs of frontal half-body models and upper-body garments, and Dress Code~\cite{morelli2022dress}, with 53,792 pairs ($1024 \times 768$) of full-body models and upper-body, lower-body, and dress garments. We focus exclusively on upper-body garments, using 11,552 training and 1,990 test pairs for VITON-HD, and 13,563 training and 1,800 test upper-body pairs for Dress Code. Dataset cleaning details are in~\cref{app:dataset_cleaning}.

\noindent \textbf{Implementation Details.}
TryOffDiff builds on pretrained Stable Diffusion v1.4~\cite{rombach2022high}, finetuning the denoising U-Net and training adapter layers from scratch. Input images are padded to a square aspect ratio and resized to $512 \times 512$ for SigLIP-B/16-512 and VAE compatibility. The adapter reduces SigLIP’s 1024 token embeddings (dimension 768) to 77 embeddings (dimension 768). Training uses a single NVIDIA A40 GPU for 150k iterations with MSE loss. Detailed parameters are in~\cref{app:training_details}.

\noindent \textbf{Baseline Approaches.}
We adapt state-of-the-art methods from pose transfer, view synthesis, and virtual try-on as baselines, modifying each to approximate garment reconstruction functionality as closely as possible.
In addition, we include IP-Adapter~\cite{ye2023ipadapter}, a general-purpose image-conditioning method, to evaluate diffusion-based image-guided generation in the VTOFF setting.
All outputs are post-processed with Segment Anything (SAM)~\cite{kirillov2023segment} using point prompts to isolate garments, then composited onto a white background, standardizing outputs and mitigating background-related artifacts that negatively impact evaluation metrics.
These baseline approaches are illustrated in~\cref{fig:baselines} and detailed in~\cref{app:baselines}.

\begin{table}[b]
\begin{center}
    \resizebox{\linewidth}{!}{
    \begin{tabular}{lcccccccc}
    \toprule
    \textbf{Method} & 
        SSIM$\uparrow$ & MS$^\text{SSIM}\uparrow$ & CW$^\text{SSIM}\uparrow$ & LPIPS$\downarrow$ & FID$\downarrow$ & FD$^\text{CLIP}\downarrow$ & KID$\downarrow$ & DISTS$\downarrow$ \\ \toprule
            GAN-Pose~\cite{roy2023multi}        & \underline{77.4} & \underline{63.8} & \underline{32.5} & \underline{44.2} & 73.2 & 30.9 & 55.8 & 30.4 \\ 
            ViscoNet~\cite{cheong2024visconet}  & 58.5 & 50.7 & 28.9 & 54.0 & 42.3 & 12.1 & 25.5 & 31.2 \\
            OOTDiff.~\cite{xu2024ootdiffusion}  & 65.1 & 50.6 & 26.1 & 49.5 & 54.0 & 17.5 & 33.2 & 32.4 \\ 
            CatVTON~\cite{chong2024catvton}     & 72.8 & 56.9 & 32.0 & 45.9 & \underline{31.4} & \underline{9.7} & \underline{17.8} & \underline{28.2} \\
            IP-Adapter~\cite{ye2023ipadapter}   & 63.2 & 53.3 & 30.1 & 51.5 & 36.1 & 13.1 & 21.6 & 30.9 \\ 
            TryOffDiff                          & \textbf{80.3} & \textbf{72.5} & \textbf{49.2} & \textbf{30.1} & \textbf{14.5} & \textbf{5.7} & \textbf{3.9} & \textbf{20.3} \\ \bottomrule
    \end{tabular}
    }
\end{center}
\caption{\textbf{Quantitative comparison on \emph{VITON-HD-test}.} 
Performance of TryOffDiff and baseline methods on the VTOFF task, evaluated using standard similarity and perceptual metrics. TryOffDiff outperforms all baselines across the board, achieving the best results on both full-reference (SSIM variants, LPIPS) and perceptual (FID, KID, DISTS) measures.
}
\label{tab:eval_metrics}
\end{table}

\noindent \textbf{Quantitative Results.}
Quantitative results on VITON-HD are shown in~\cref{tab:eval_metrics}. TryOffDiff outperforms baselines across all metrics, achieving a DISTS score of 20.3, a 7.9-point improvement over CatVTON (28.2). Baseline rankings vary: GAN-Pose excels in full-reference metrics (\eg, SSIM 77.4), while CatVTON leads in no-reference metrics (\eg, FID 31.4). DISTS, balancing structure and texture, aligns with visual perception, ranking CatVTON highest among baselines.

We examine the robustness of our approach by evaluating its cross-dataset generalization. On Dress Code, TryOffDiff achieves a DISTS score of 21.6 for upper-body garments in the within-dataset setting (DC/DC, \cref{tab:eval_metrics_full}). Baselines were not evaluated on Dress Code due to their poor performance in initial tests. As our main focus is on upper-body garments, results for lower-body garments and dresses are provided in~\cref{app:supp_results} for completeness.
Importantly, in the cross-dataset setting, TryOffDiff trained on Dress Code and tested on VITON-HD (DC/HD) outperforms all VITON-HD baselines across all metrics (\eg, DISTS 26.6 vs. CatVTON’s 28.2), despite domain shifts (\eg, full-body to half-body references). This demonstrates the model’s strong generalization capabilities, although performance remains below the in-domain HD/HD setting (DISTS 20.3), which is expected under domain shift.

\begin{table}[t]
\begin{center}
    \resizebox{\linewidth}{!}{
    \begin{tabular}{lcccccccc}
    \toprule
    \textbf{Train/Test} & 
        SSIM$\uparrow$ & MS$^\text{SSIM}\uparrow$ & CW$^\text{SSIM}\uparrow$ & LPIPS$\downarrow$ & FID$\downarrow$ & FD$^\text{CLIP}\downarrow$ & KID$\downarrow$ & DISTS$\downarrow$ \\ \toprule
            HD\,/\,HD & 80.3 & 72.5 & 49.2 & 30.1 & 14.5 & 5.7 & 3.9 & 20.3 \\
            DC\,/\,HD & 77.8 & 65.5 & 37.0 & 38.3 & 22.9 & 7.9 & 9.2 & 26.6 \\ \midrule 
            DC\,/\,DC & 80.8 & 73.8 & 47.8 & 31.6 & 17.1 & 5.2 & 4.7 & 21.6 \\
            HD\,/\,DC & 75.8 & 65.4 & 34.7 & 41.5 & 29.6 & 9.7 & 10.5 & 28.1 \\ \bottomrule            
    \end{tabular}
    }
\end{center}
\caption{\textbf{Within- and cross-dataset evaluation of TryOffDiff.}
Metrics for TryOffDiff trained and tested on VITON-HD (HD) or DressCode (DC, upper-body) in within-dataset (HD/HD, DC/DC) and cross-dataset (DC/HD, HD/DC) settings.}
\label{tab:eval_metrics_full}
\end{table}

\begin{figure}[t]
\begin{center}
  \subcaptionbox{\centering \footnotesize Reference}{\includegraphics[width=.115\linewidth]{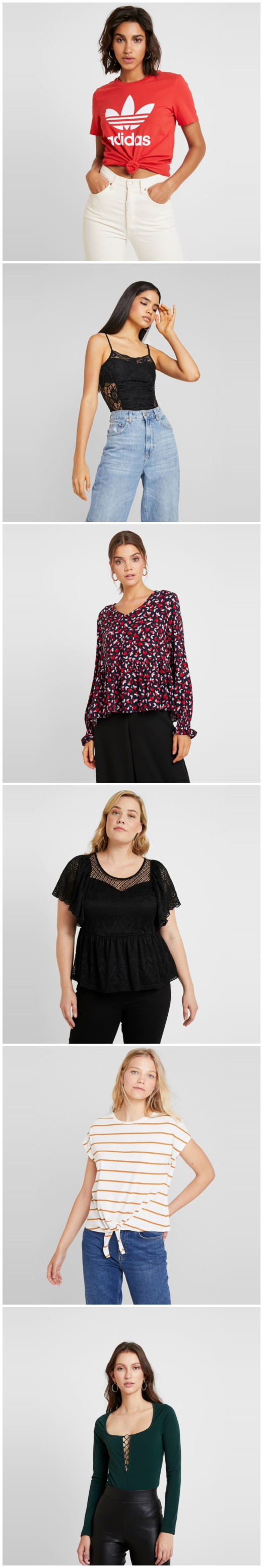}} \hfill
  \subcaptionbox{\centering \footnotesize Gan-Pose}{\includegraphics[width=.115\linewidth]{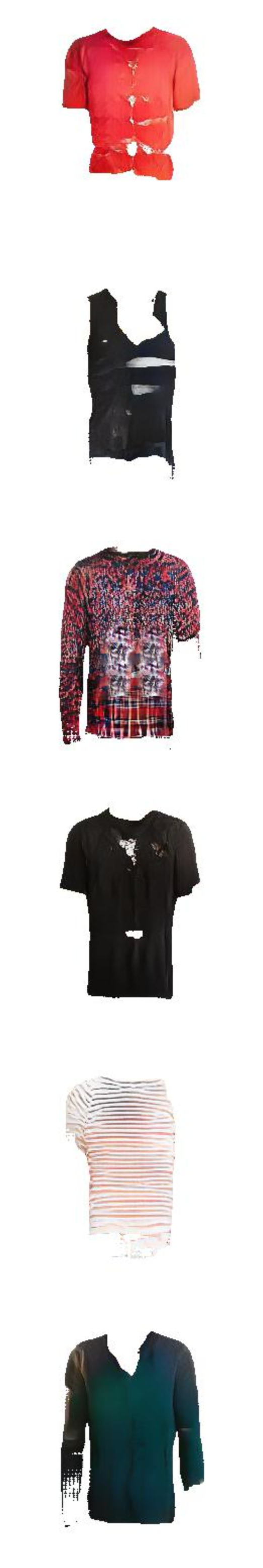}} \hfill
  \subcaptionbox{\centering \footnotesize ViscoNet}{\includegraphics[width=.115\linewidth]{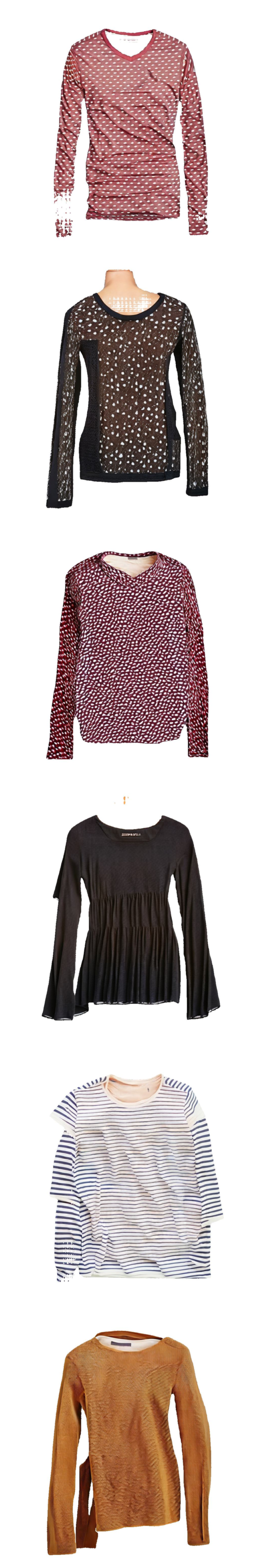}} \hfill
  \subcaptionbox{\centering \footnotesize OOTDiff.}{\includegraphics[width=.115\linewidth]{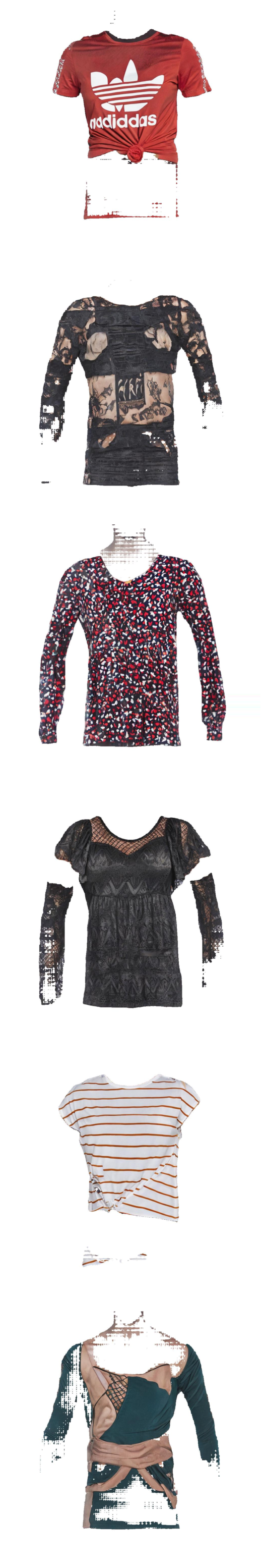}} \hfill
  \subcaptionbox{\centering \footnotesize CatVTON}{\includegraphics[width=.115\linewidth]{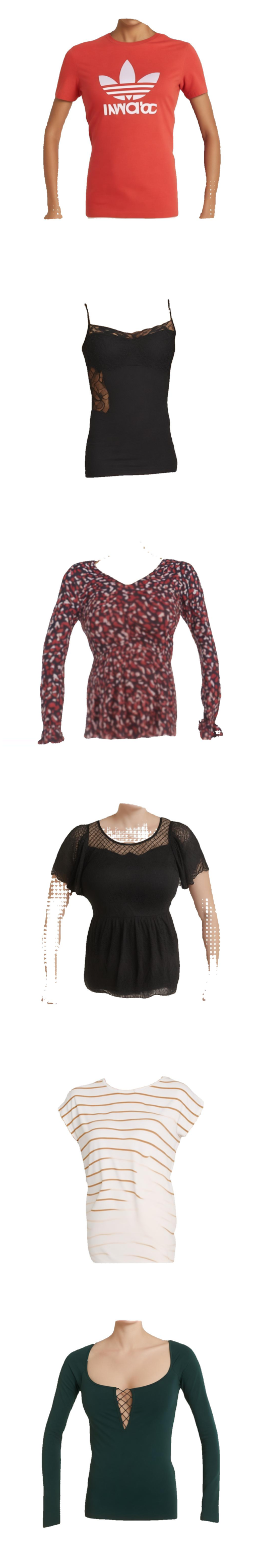}} \hfill
  \subcaptionbox{\centering \footnotesize IP-Adapter}{\includegraphics[width=.115\linewidth]{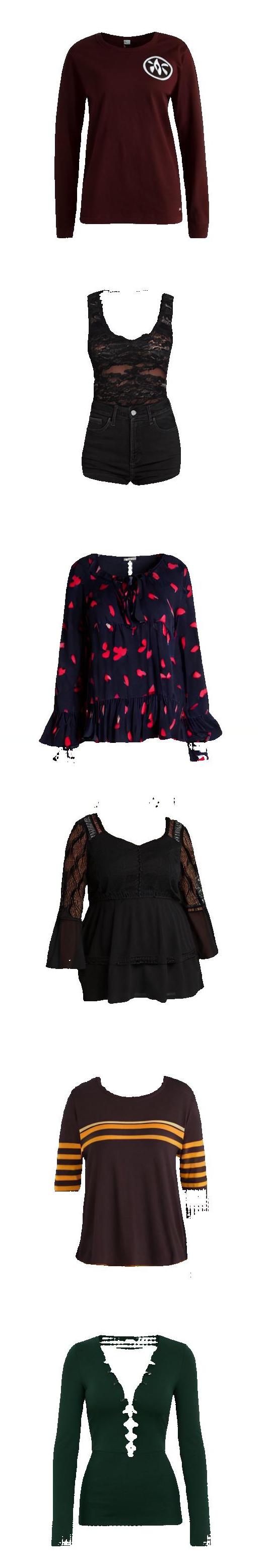}} \hfill  
  \subcaptionbox{\centering \footnotesize TryOffDiff}{\includegraphics[width=.115\linewidth]{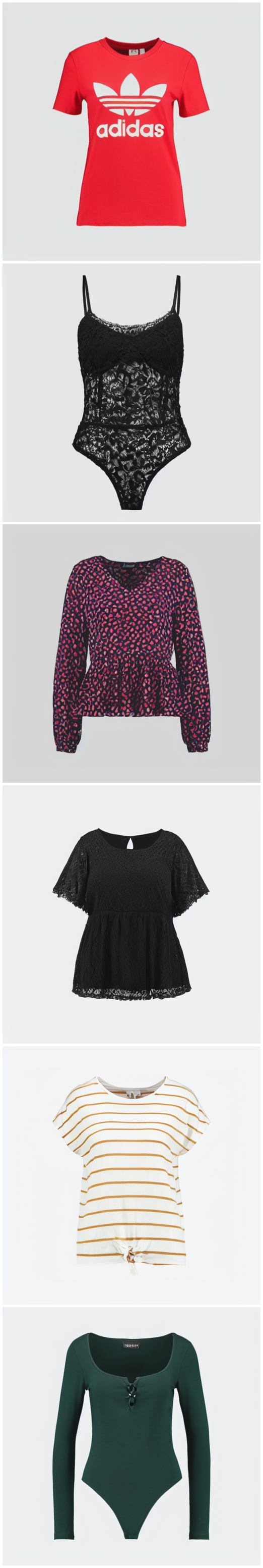}} \hfill
  \subcaptionbox{\centering \footnotesize Target}{\includegraphics[width=.115\linewidth]{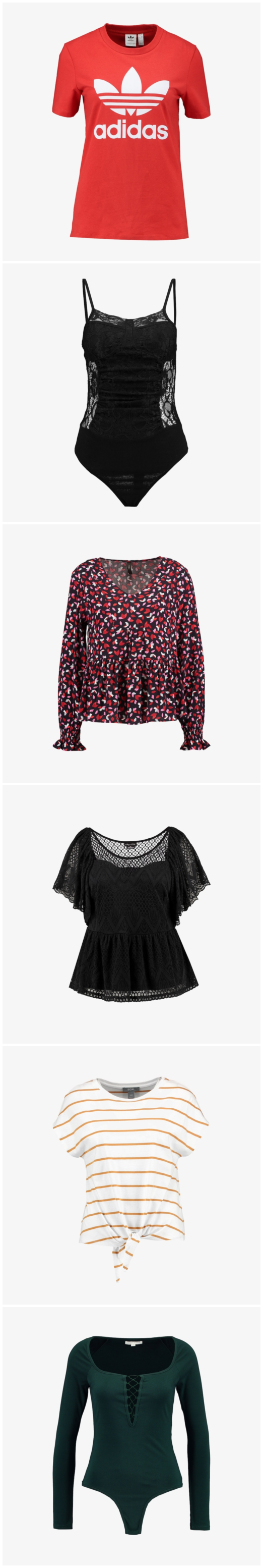}}
\end{center}  
  \caption{\textbf{Qualitative comparison on \emph{VITON-HD-test}.} In comparison to the baseline approaches, TryOffDiff is capable of generating garment images with accurate structural details as well as fine textural details.
  }
  \label{fig:comparison1}
\end{figure}

\noindent \textbf{Qualitative Analysis.}
Qualitative results for VITON-HD are shown in both \cref{fig:cover} and \cref{fig:comparison1}, while Dress Code examples appear in \cref{fig:cover} (more in \cref{app:supp_results}).
These examples align with our quantitative findings and highlight metric-specific strengths, as discussed in~\cref{sec:vtoff}. GAN-Pose approximates garment color and shape but often misses small regions, reducing visual fidelity in no-reference metrics. ViscoNet produces realistic outputs but deforms shapes and favors long sleeves, with limited textural detail. OOTDiffusion preserves logos but struggles to maintain texture consistency. CatVTON excels at rendering natural textures and logos but often mismatches garment shapes, limiting full-reference performance.
In contrast, TryOffDiff consistently reconstructs accurate shapes and textures, including occluded regions (\eg, high-cut bodysuits in Dress Code), colors, patterns, and logos, achieving superior DISTS scores. Its image conditioning mechanism enables precise recovery, outperforming baselines. 
Additional qualitative results
are provided in~\cref{app:supp_results}.

\noindent \textbf{Ablations.}
We ablate TryOffDiff’s components on VITON-HD (\Cref{tab:ablations_vitonhd}). Using SigLIP-B/16 over CLIP ViT-B/32 improves DISTS from 23.5 to 22.1 without pretraining, and to 20.3 with pretraining, due to enhanced feature extraction. The adapter (Linear+LN) boosts performance (\eg, DISTS 23.7 to 23.5 with CLIP), and pretraining Stable Diffusion v1.4 further improves all metrics (\eg, DISTS 22.1 to 20.3). These results confirm the importance of SigLIP and pretraining for high-fidelity garment reconstruction. Additional ablation analyses are in~\cref{app:supp_results}.

\begin{table}[b]
\begin{center}
\resizebox{\linewidth}{!}{
\begin{tabular}{cccc|ccc|cccc|c}
    \toprule
    \multicolumn{4}{c|}{Training Configuration} & \multicolumn{3}{c|}{Structural Similarity$\uparrow$} & \multicolumn{4}{c|}{Perceptual Quality$\downarrow$} & \\
    Img. Encoder & Adapter & Cond. Shape & Pretrain & SSIM & MS$^\text{SSIM}$ & CW$^\text{SSIM}$ & LPIPS & FID & FD$^\text{CLIP}$ & KID & DISTS$\downarrow$ \\
    \midrule
     CLIP ViT-B/32 & -         & (257,1024) & - & 77.9 & 69.2 & 46.4 & 35.3 & 16.8 & 7.7 & 5.2 & 23.7 \\  
     CLIP ViT-B/32 & Linear+LN & (77,768) & - & 78.8 & 70.5 & 46.5 & 34.3 & 16.3 & 7.4 & 4.7 & 23.5 \\  
     SigLIP-B/16   & Linear+LN & (77,768) & - & 78.4 & 69.8 & 47.3 & 33.1 & 16.4 & 6.6 & 5.1 & 22.1 \\  
     CLIP ViT-B/32 & Linear+LN & (77,768) & \checkmark & 79.8 & 71.5 & 47.8 & 32.0 & 15.2 & 5.8 & 4.1 & 21.5 \\  
     SigLIP-B/16   & Linear+LN & (77,768) & \checkmark & \textbf{80.3} & \textbf{72.5} & \textbf{49.2} & \textbf{30.1} & \textbf{14.5} & \textbf{5.7} & \textbf{3.9} & \textbf{20.3} \\  
    \bottomrule
\end{tabular}
}
\end{center}
\caption{\textbf{Ablation study on VITON-HD.} 
Training configurations and metrics on \emph{VITON-HD-test}.
``Pretrain" indicates U-Net initialization from Stable Diffusion (\checkmark) or scratch (-). 
All ablations mirror Stable Diffusion-v1.4 architecture.}
\label{tab:ablations_vitonhd}
\end{table}

\section{Conclusion}\label{sec:conclusion}

In this paper, we introduced virtual try-off (VTOFF), a novel task designed to reconstruct standardized upper-body garment images from a single reference photo of a person wearing the garment. Unlike virtual try-on (VTON), VTOFF emphasizes precise garment reconstruction, making it ideal for evaluating generative models’ accuracy in capturing fine details such as patterns and logos.
We proposed TryOffDiff, an adaptation of Stable Diffusion with SigLIP image conditioning to guide the generative process, replacing text-based inputs. Evaluated on VITON-HD and Dress Code (upper-body garments), it outperforms baselines adapted from pose transfer and VTON methods, achieving the best performance across all evaluation metrics. Cross-dataset experiments further demonstrate its robust generalization despite domain shifts (\eg, full-body to half-body references). Qualitative results confirm TryOffDiff's strength in preserving garment shape, texture, and occluded regions, with minimal pre- and post-processing.
While TryOffDiff performs well overall, challenges remain in reconstructing garments with highly complex structures, such as intricate logos and text. Looking ahead, VTOFF can be extended to multi-garment scenarios and enriched with more advanced conditioning or generation architectures, offering new avenues for evaluating generative models and advancing applications in e-commerce.

\section*{Acknowledgment}
{This work has been funded by Horizon Europe program under grant agreement 101134447-ENFORCE, and by the German federal state of North Rhine-Westphalia as part of the research funding program KI-Starter. We would like to thank UniZG-FER for providing access to their hardware.}

\bibliography{egbib}

\appendix
\clearpage
\setcounter{page}{1}
\maketitlesupplementary

\section{Dataset Cleaning}\label{app:dataset_cleaning}
Upon closer inspection of VITON-HD, we identified 95 duplicate image pairs (0.8\%) in the training set and 6 duplicate pairs (0.3\%) in the test set. Additionally, we found 36 pairs (1.8\%) in the training set that had been included in the original test split. To ensure the integrity of our experiments, we cleaned the dataset by removing all duplicates in both subsets as well as all leaked examples from the test set. The resulting cleaned dataset, contains 11,552 unique image pairs for training and 1,990 unique image pairs for testing. We provide the script for cleaning the dataset in our code repository.

\section{Training Details}\label{app:training_details}
We train TryOffDiff by building on the
pretrained Stable Diffusion v1.4~\cite{rombach2022high},
focusing on finetuning the denoising U-Net
and training adapter layers from scratch.
As a preprocessing step,
we pad the input reference image along the width for a square aspect ratio,
then resize them to a resolution of $512 \times 512$
to match the expected input format
of the pretrained SigLIP and VAE encoder.
For training, we preprocess the garment images in the same way.
We use SigLIP-B/16-512 as image feature extractor, 
which outputs 1024
token embeddings of dimension 768.
Our adapter, consisting of linear and normalization layers, 
reduces these to $n=77$ conditioning embeddings of dimension $m=768$.

Training occurs over 150k iterations 
on a single NVIDIA A40 GPU, 
requiring approximately 5 days with a batch size of 16. 
We employ the AdamW optimizer~\cite{loshchilov2018decoupled}, 
with an initial learning rate of 1e-4
that increases linearly from 0 during the first \%5 warmup steps, 
then follows a linear decay to 0. 
As proposed in~\cite{karras2022elucidating}, 
we use the EulerDiscrete scheduler with 1,000 steps.
We optimize using the standard Mean Squared Error (MSE) loss, 
which measures the difference between the added 
and the predicted noise at each step.
This loss function is commonly employed 
in diffusion models to guide the model 
in learning to reverse the noising process effectively. 
During inference, we run TryOffDiff 
with a EulerDiscrete scheduler over 20 timesteps with a guidance scale of 2.0.
On a single NVIDIA A40 GPU, 
this process takes 1.8 seconds per image and requires 9.8GB of memory.

\section{Baselines}\label{app:baselines}

\noindent \textbf{GAN-Pose}~\cite{roy2023multi}
is a GAN-based pose transfer method that
expects three inputs: a reference image, and pose heatmaps of the reference and target subject.
Garment images from VITON-HD are used to estimate the
heatmap for a fixed, neutral pose.
This setup enables the transfer of human poses
from diverse reference images to a standardized pose,
aligning the output to the typical view of product images.

\noindent \textbf{ViscoNet}~\cite{cheong2024visconet}
requires a text prompt, a pose, a mask, 
and multiple masked conditioning images as inputs. 
For the text prompt, we use a description such as ``a photo of an e-commerce clothing product''.
We choose a garment image from VITON-HD to 
estimate a neutral pose as well as a 
generic target mask.
Since ViscoNet is originally trained with masked conditioning images,
we apply an off-the-shelf fashion parser~\cite{velioglu2024fashionfail} 
to mask the upper-body garment, which is then provided as input.

\noindent \textbf{OOTDiffusion}~\cite{xu2024ootdiffusion}
takes a garment image and a reference image to generate 
a VTON output. 
To adapt this model for VTOFF,
we again apply the fashion parser~\cite{velioglu2024fashionfail} 
to mask the upper-body garment to create the garment image.
We select a reference image with a mannequin 
in a neutral pose as further input. 
An intermediate step involves masking
the upper-body within the reference image, 
for which we use a hand-crafted masked
version of the reference image. 

\noindent \textbf{CatVTON}~\cite{chong2024catvton}
is a model that generates a VTON image
using a reference image and a conditioning garment
image as inputs. An intermediate step 
incorporates upper-body masks to guide the try-on process. 
For adaptation to VTOFF, 
we replace the reference image with a plain white image 
and use a handcrafted mask in a neutral pose, 
enabling CatVTON to perform garment transfer
independently of any specific person.

\noindent \textbf{IP-Adapter}~\cite{ye2023ipadapter}
integrates into Stable Diffusion-v1.5 without architecturla changes, enabling image-based conditioning via image prompts.
For VTOFF, we adapt IP-Adapter to reconstruct standardized garment images using visual cues from a reference image.
The generation is initialized from a mask-like image on a white background, serving as a neutral garment template. A reference image of a clothed person guides the synthesis, with a generic text prompt (``high quality product photo of a garment'') and a high image prompt weight (0.75) to ensure fidelity to garment details while maintaining a consistent, catalog-style presentation.

\section{Implementation Details}
For evaluation, we use `IQA-PyTorch'~\cite{pyiqa} to compute SSIM, MS-SSIM, CW-SSIM, and LPIPS, and the `clean-fid'~\cite{parmar2022aliased} library for FID, CLIP-FID, and KID. We employ the original implementation of DISTS~\cite{ding2020image} for evaluating perceptual image quality. For readability purposes, the values of SSIM, MS-SSIM, CW-SSIM, LPIPS, and DISTS presented in this paper are multiplied by 100, and KID is multiplied by 1000.

\section{Additional Results}\label{app:supp_results}

To further validate DISTS's robustness over traditional metrics (as discussed in~\cref{sec:vtoff}), we conducted experiments on the entire \emph{VITON-HD-test} dataset, with averaged results reported in~\cref{app:tab:eval_metrics_test}. 
The results confirm that widely used metrics for VTON, such as SSIM and its variants, fail to capture image similarity in a manner suitable for VTON and VTOFF.
For instance, SSIM-based metrics assign the highest score to case (b)—a structurally correct but color-distorted image—while LPIPS similarly overestimates similarity in this case. In contrast, DISTS assigns its lowest score to case (e), which is arguably the most perceptually similar to the ground truth for garment-focused tasks. These findings highlight the inadequacy of traditional metrics for both VTON and VTOFF and establish DISTS as a more reliable measure for evaluating garment image similarity.

\begin{table}
\begin{center}
\begin{tabular}{c|ccccc}
    \toprule
    \textbf{Case} & 
        SSIM$\uparrow$ & MS$^\text{SSIM}\uparrow$ & CW$^\text{SSIM}\uparrow$ & LPIPS$\downarrow$ & DISTS$\downarrow$ \\ \toprule
        (a) & 82.59    & 71.31             & 40.31             & 35.05             & 21.43 \\
        (b) & 98.81    & 98.85             & 96.42             & 12.14             & 14.43 \\
        (c) & 90.24    & 78.72             & 58.91             & 33.77             & 19.76 \\ \midrule
        (d) & 73.08    & 66.09             &  0                & 45.65             & 69.54  \\
        (e) & 81.48    & 77.15             & 72.35             & 17.42             & 10.04 \\
        (f) & 77.52    & 92.63             & 73.41             & 20.91             & 29.85 \\
            \bottomrule
\end{tabular}
\end{center}
\caption{\textbf{Quantitative evaluation of metric suitability.} 
Average values of various metrics for different scenarios, as shown in \cref{fig:metrics_failures}. Metrics are calculated at a full resolution of 1024x768 using the \emph{VITON-HD-test} dataset.
}
\label{app:tab:eval_metrics_test}
\end{table}

\Cref{app:tab:dc_results} presents evaluation metrics for TryOffDiff models, each trained separately on a distinct garment category from the Dress Code dataset: upper-body, lower-body, and dresses. The results demonstrate robust performance across all categories, with structural metrics (SSIM, MS-SSIM, CW-SSIM) and perceptual metrics (DISTS, LPIPS) showing minimal variation. Dresses exhibit slightly improved DISTS and LPIPS scores, indicating that TryOffDiff effectively handles garments with complex global structures. These findings highlight the model’s adaptability and effectiveness across diverse garment types.

\begin{table}
\begin{center}
    \resizebox{\linewidth}{!}{
    \begin{tabular}{lcccccccc}
    \toprule
    \textbf{Garment Type} & 
        SSIM$\uparrow$ & MS$^\text{SSIM}\uparrow$ & CW$^\text{SSIM}\uparrow$ & LPIPS$\downarrow$ & FID$\downarrow$ & FD$^\text{CLIP}\downarrow$ & KID$\downarrow$ & DISTS$\downarrow$ \\ \toprule
            Upper-body & 80.8 & 73.8 & 47.8 & 31.6 & 17.1 & 5.2 & 4.7 & 21.6 \\
            Lower-body & 81.1 & 76.8 & 55.5 & 30.0 & 22.6 & 9.4 & 5.9 & 21.1 \\
            Dresses    & 81.6 & 76.1 & 55.6 & 26.1 & 18.4 & 8.2 & 4.4 & 20.8 \\
            \bottomrule            
    \end{tabular}
    }
\end{center}
\caption{
\textbf{Evaluation metrics for TryOffDiff models by garment type.} 
Models are trained independently on upper-body, lower-body, and dress categories from Dress Code dataset.}
\label{app:tab:dc_results}
\end{table}

\begin{figure}
  \centering
  \begin{subfigure}{0.15\linewidth}
    \includegraphics[width=\linewidth]{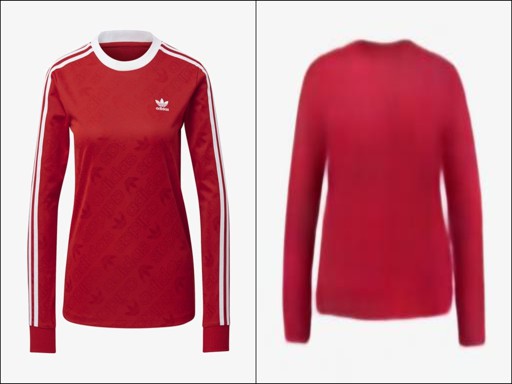}
    \caption{$81.9 \text{ / } 36.2$}
  \end{subfigure}
  \hfill
  \begin{subfigure}{0.15\linewidth}
    \includegraphics[width=\linewidth]{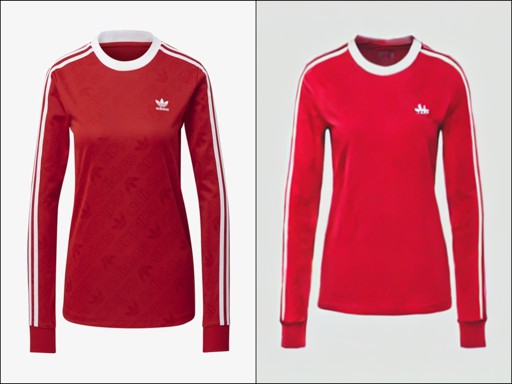}
    \caption{$80.3 \text{ / } 24.2$}
  \end{subfigure}
  \hfill  
  \begin{subfigure}{0.15\linewidth}
    \includegraphics[width=\linewidth]{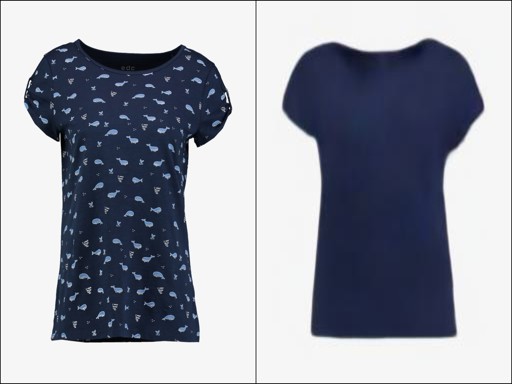}
    \caption{$81.5 \text{ / } 40.4$}
  \end{subfigure}
  \hfill
  \begin{subfigure}{0.15\linewidth}
    \includegraphics[width=\linewidth]{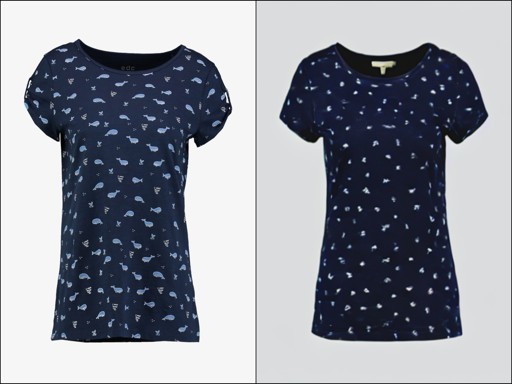}
    \caption{$75.3 \text{ / } 25.0$}
  \end{subfigure}  
  \hfill  
  \begin{subfigure}{0.15\linewidth}
    \includegraphics[width=\linewidth]{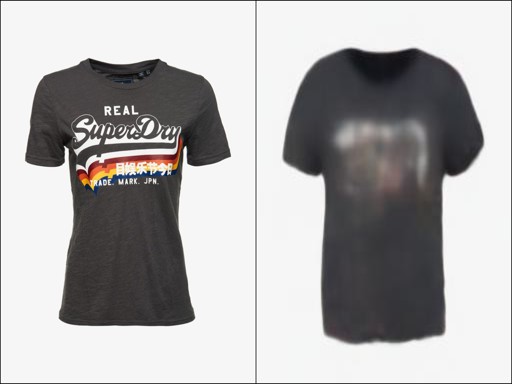}
    \caption{$81.7 \text{ / } 39.7$}
  \end{subfigure}
  \hfill  
  \begin{subfigure}{0.15\linewidth}
    \includegraphics[width=\linewidth]{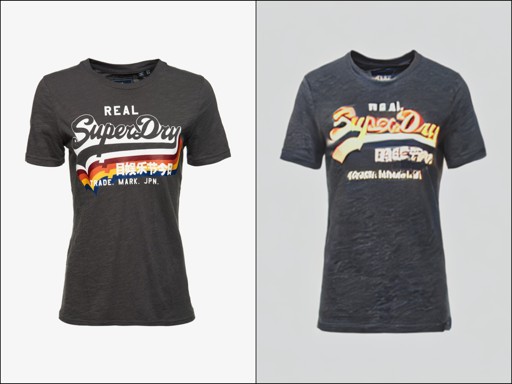}
    \caption{$80.3 \text{ / } 19.4$}
  \end{subfigure}  
  \\
  \caption{\textbf{Metric suitability for VTOFF (SSIM\textuparrow ~/~DISTS\textdownarrow).} 
  Each pair displays ground truth (left) and predictions (right) from two arbitrary models (a-c vs. d-f). High SSIM scores obscure poor quality in the top row, while DISTS effectively highlights variations, better reflecting human perception.
  }
  \label{fig:metric_failures_2}
\end{figure}

To further demonstrate the limitations of the widely used SSIM metric in both VTON and VTOFF tasks, we compare ground-truth images with predictions from two arbitrary models, as shown in \cref{fig:metric_failures_2}. The results highlight that SSIM often fails to penalize perceptual artifacts. For example, in \cref{fig:metric_failures_2}(c), the prediction is blurry and lacks detail, yet receives a higher SSIM score than \cref{fig:metric_failures_2}(d), which preserves patterns and offers a more faithful reconstruction. Similar issues are observed across other examples. In contrast, DISTS more reliably reflects perceptual quality, assigning lower scores to visually degraded outputs and aligning better with human judgment.

\begin{figure}[t]
\begin{center}
  \subcaptionbox{\centering \footnotesize Reference}{\includegraphics[width=0.115\linewidth]{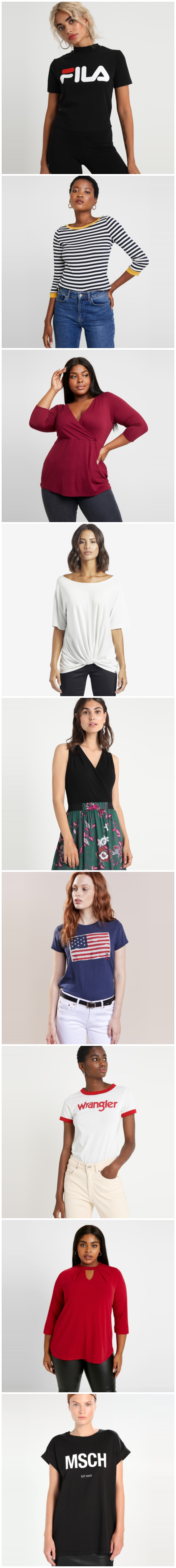}} \hfill
  \subcaptionbox{\centering \footnotesize Gan-Pose}{\includegraphics[width=0.115\linewidth]{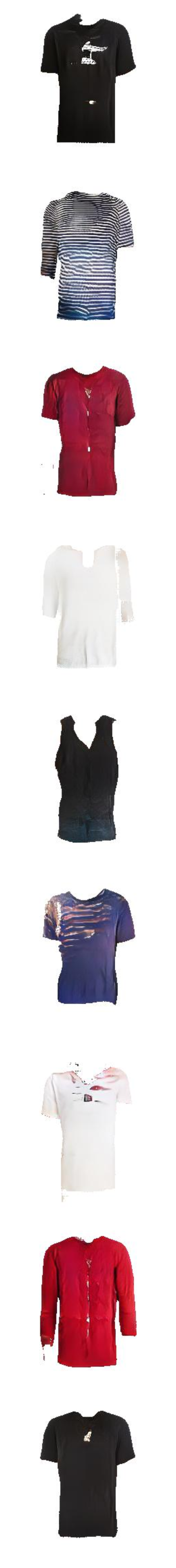}} \hfill
  \subcaptionbox{\centering \footnotesize ViscoNet}{\includegraphics[width=0.115\linewidth]{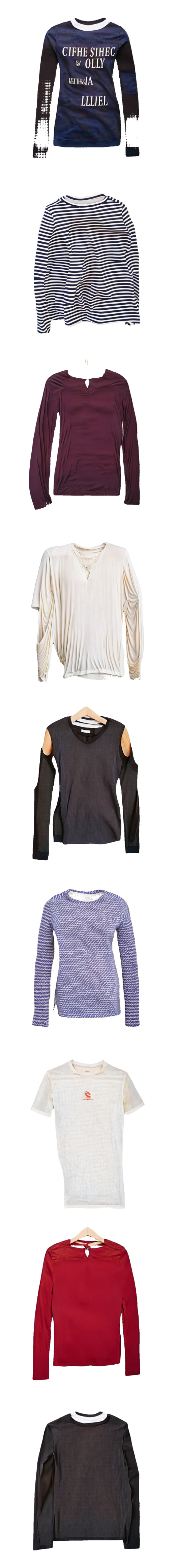}} \hfill
  \subcaptionbox{\centering \footnotesize OOTDiff.}{\includegraphics[width=0.115\linewidth]{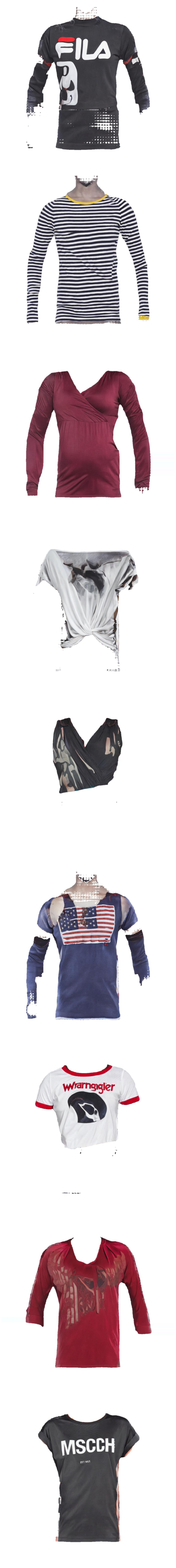}} \hfill
  \subcaptionbox{\centering \footnotesize CatVTON}{\includegraphics[width=0.115\linewidth]{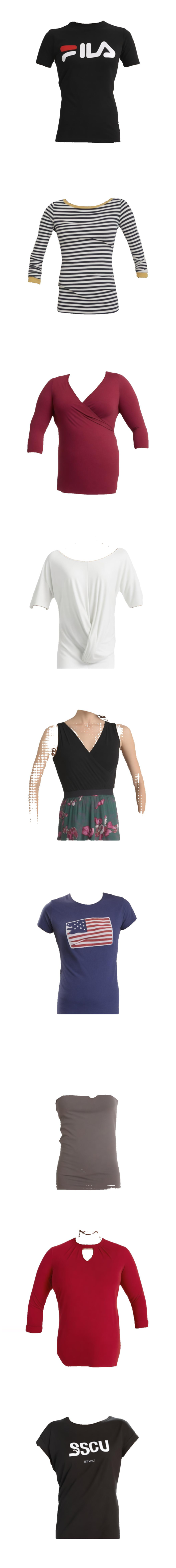}} \hfill
  \subcaptionbox{\centering \footnotesize IP-Adapter}{\includegraphics[width=0.115\linewidth]{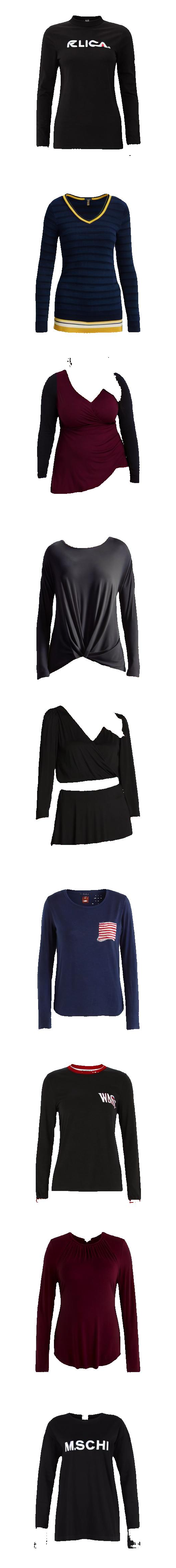}} \hfill  
  \subcaptionbox{\centering \footnotesize TryOffDiff}{\includegraphics[width=0.115\linewidth]{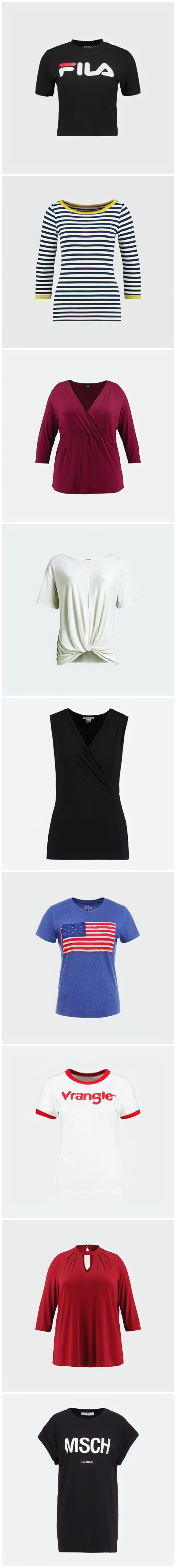}} \hfill
  \subcaptionbox{\centering \footnotesize Target}{\includegraphics[width=0.115\linewidth]{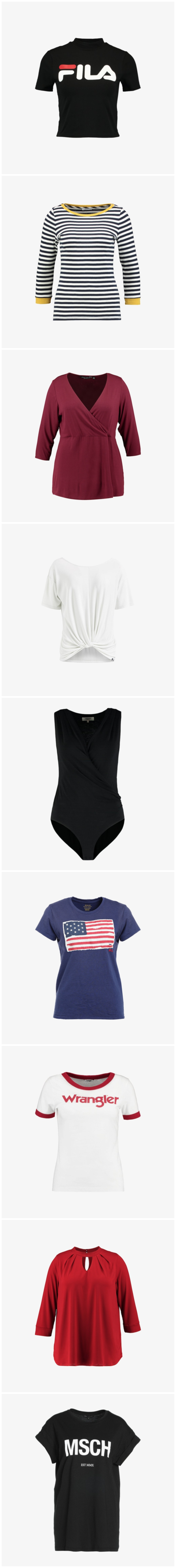}}
\end{center}  
  \caption{\textbf{Qualitative comparison between baselines and TryOffDiff on VITON-HD.} TryOffDiff more accurately reconstructs both structural and textural garment details compared to baseline methods.}
\label{suppl-fig:comparison}
\end{figure}

\begin{figure}[t]
\begin{center}
    \includegraphics[width=\linewidth]{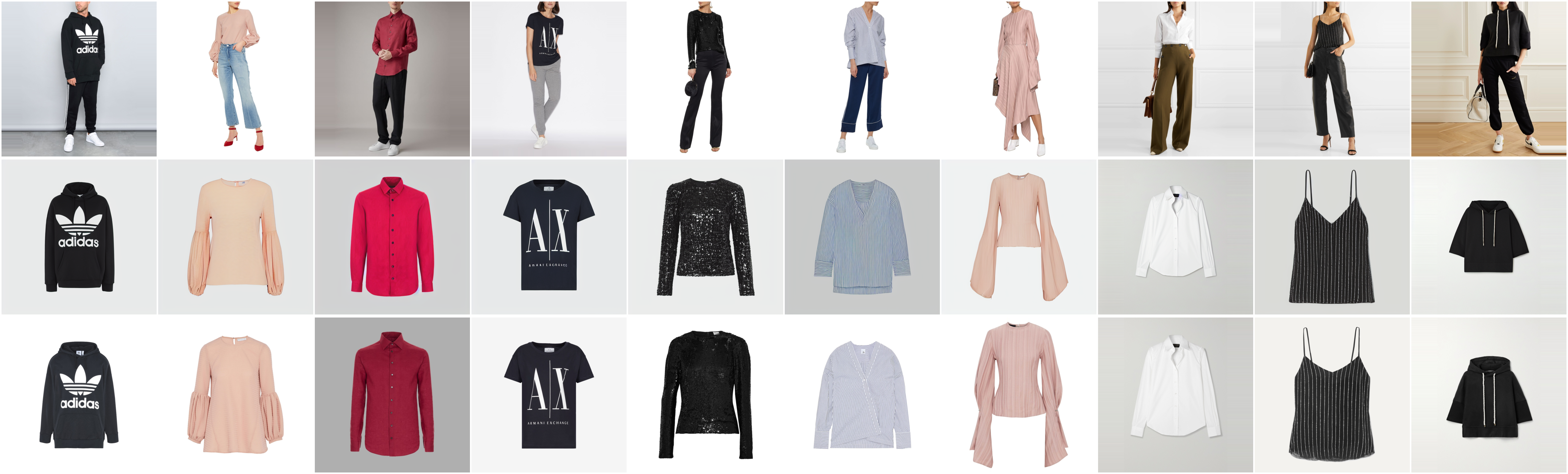}
\vspace{1em}
    \includegraphics[width=\linewidth]{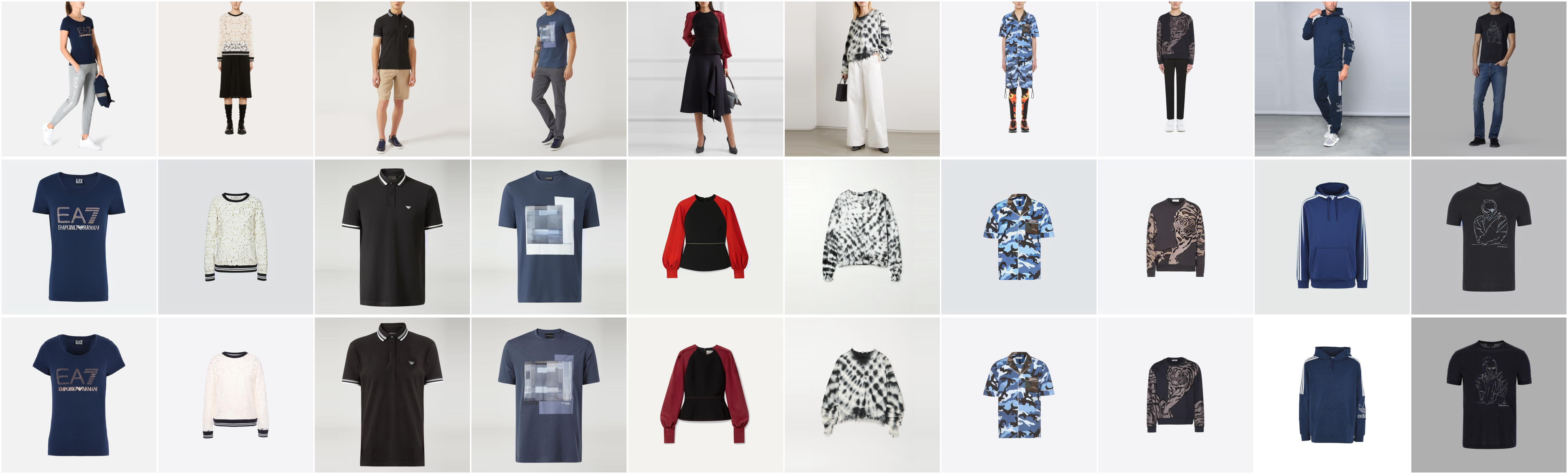}
\vspace{1em}
  \includegraphics[width=\linewidth]{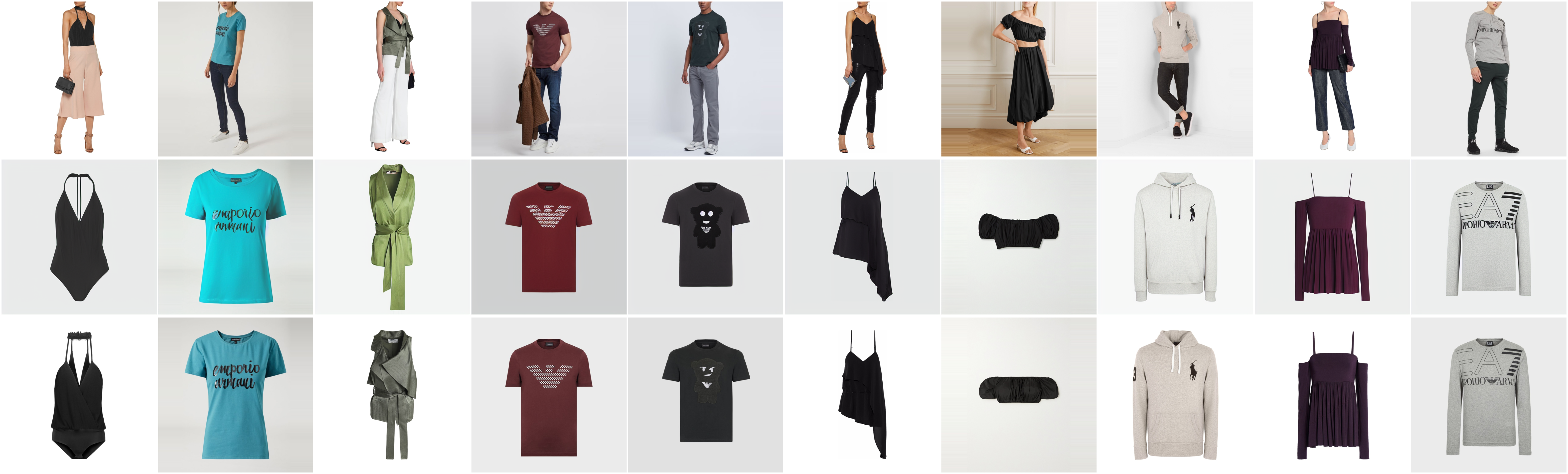}
\end{center}  
\caption{
\textbf{Qualitative results from TryOffDiff on Dress Code (upper-body garments).}
}
\label{suppl-fig:dc_samples}
\end{figure}

\end{document}